\documentclass[runningheads,a4paper]{llncs}

\usepackage{amssymb}
\usepackage{graphicx}
\usepackage{float}

\usepackage{url}
\urldef{\mailsa}\path|{alfred.hofmann, ursula.barth, ingrid.haas, frank.holzwarth,|
	\urldef{\mailsb}\path|anna.kramer, leonie.kunz, christine.reiss, nicole.sator,|
	\urldef{\mailsc}\path|erika.siebert-cole, peter.strasser, lncs}@springer.com|  
\newcommand{\keywords}[1]{\par\addvspace\baselineskip
	\noindent\keywordname\enspace\ignorespaces#1}

\floatstyle{ruled}
\newfloat{algorithm}{tbp}{loa}
\providecommand{\algorithmname}{Algorithm}

\floatname{algorithm}{\protect\algorithmname}
\usepackage{amsmath,graphicx}
\usepackage{epsfig}
\usepackage{graphics} 
\usepackage{epsfig} 
\usepackage{amsmath}
\usepackage{url}
\usepackage{tablefootnote}
\usepackage{subfig}

\usepackage{makecell}
\providecommand{\tabularnewline}{\\}

\begin{document}
	
	\mainmatter  
	
	\title{Deep multiscale convolutional feature learning for weakly supervised localization of chest pathologies in X-ray images}
	
	\titlerunning{ Weakly supervised Localization of chest pathologies}
	%
	%
	%
	

\author{Suman Sedai, Dwarikanath Mahapatra, Zongyuan Ge,Rajib Chakravorty and Rahil Garnavi}
\institute{IBM Research - Australia, Melbourne, VIC, Australia, \\ 
	\email{ssedai@au1.ibm.com}
}

	%
	%
	
	\toctitle{Lecture Notes in Computer Science}
	\tocauthor{Authors' Instructions}
	\maketitle

	\begin{abstract}
		
		Localization of chest pathologies in chest X-ray images is a challenging task because of their varying sizes and appearances. We propose a novel weakly supervised method to localize chest pathologies using class aware deep multiscale feature learning. Our method leverages intermediate feature maps from CNN layers at different stages of a deep network during the training of a classification model using image level annotations of pathologies. During the training phase, a set of \emph{layer relevance weights} are learned for each pathology class and the CNN is optimized to perform pathology classification by convex combination of feature maps from both shallow and deep layers using the learned weights. During the test phase, to localize the predicted pathology, the multiscale attention map  is obtained by convex combination of class activation maps from  each stage using the \emph{layer relevance weights} learned during the training phase. We have validated our method using 112000 X-ray images and compared with the state-of-the-art localization methods. We experimentally demonstrate that the proposed weakly supervised method can improve the localization performance of small pathologies such as nodule and mass while giving comparable performance for bigger pathologies e.g., Cardiomegaly.


		

		
		
		
	\end{abstract}

	\section{Introduction} 
	\label{sec:intro}
	Chest X-ray is very economical and the most commonly used imaging modality for screening and diagnosis of many lung diseases. There is an exponential growth in the number of X-ray images taken in hospitals that must be reviewed by radiologists. Manual examination of scans is time consuming and subjective. Therefore, automated systems that can assess chest X-ray images will greatly assist radiologists and health care centers in managing patients and critical operations. Moreover, automated localization and annotation of pathologies and disease areas within the scan and providing those visualization to radiologist would allow clinicians to better understand the system's assessment and evaluate its reliability.
	
	Existing object localization methods are based on patch classification \cite{Liao2013}, region-based convolution networks \cite{Jia2017,Girshick2015}, fully convolutional neural networks \cite{Olaf2015,Sedai2017,seg2017}.  These approaches are \emph{fully supervised} approach i.e., they require location-level annotation of object being detected during training phase. Acquiring such annotations is a tedious process and is expensive to perform over large data-sets. \emph{Weakly supervised} methods on the other hand, can predict the location of object of interest with only image level annotation in training time. Therefore, it bypasses the need for the bounding box location annotation of pathologies. In this paper, we propose a novel weakly supervised method based on Convolutional neural network (CNN) by leveraging the intermediate feature maps of CNN to localize the chest pathologies in X-ray images. 
	
	Early works on weakly supervised methods use \emph{multiple instance learning} and \emph{bag of words} for chest pathology localization in X-ray images \cite{Uri2011} and cancer cell detection in histopathology images \cite{XU2014591}. Recent work have shown that CNN trained using image level annotation alone can be used to localize the object of interest \cite{Zhou2016,Oquab2015}. The global pooling of convolution layers in CNN retains spatial information about the discriminative regions in the image which can be used to compute the class activation map (CAM) \cite{Zhou2016}. CAM gives the relative importance of the layer activation at different 2D spatial locations, can be used as saliency map to localize the object. In medical imaging domain, CAM based methods have been developed for tuberculosis detection in X-ray images \cite{Sangheum2016}. In another work, soft attention map obtained from CAM have been combined with LSTM network to detect lung nodule in chest X-ray images \cite{Pesce2017}. Recently, \cite{Wang2017} used weakly supervised method and CAM to localize the chest pathologies. These approaches, however, only use activation maps from the deepest convolution layers where the resolution of feature maps have been reduced to minimum amongst all the layers, which means localization ability of the network is dependent on the spatial resolution of the last convolution layer \cite{Zhou2016}. 
	
	
	However, using feature maps from only highest convolution layers may adversely affect the localization of small pathologies.  Successful localization of small pathologies, such as nodule, may  increase the accuracy in incidental findings during routine check-ups and, therefore, the efficacy of chest x-ray based investigation. Therefore, we propose a weakly supervised localization method based on CNN using multiscale learning of feature maps at both shallower and deeper layers.  The proposed method also learns  the layer-wise relevance  weights which determines the relative importance of each layer to classify a given pathology.  The learned layer-wise relevance information is then used to combine the feature maps from individual layers. Thus, allowing pathologies to obtain multiscale attention map from different layers according to their relevance in classification process. The main advantage of the proposed method is its ability to localize chest pathologies of different sizes, and particularly those with small sizes which are often more challenging, using weak labels (image level annotation) only.

	
	
	

	\section{Methodology} 
	
	
	
	
	
\begin{figure}[t]
		
		\centering \includegraphics[scale=0.19]{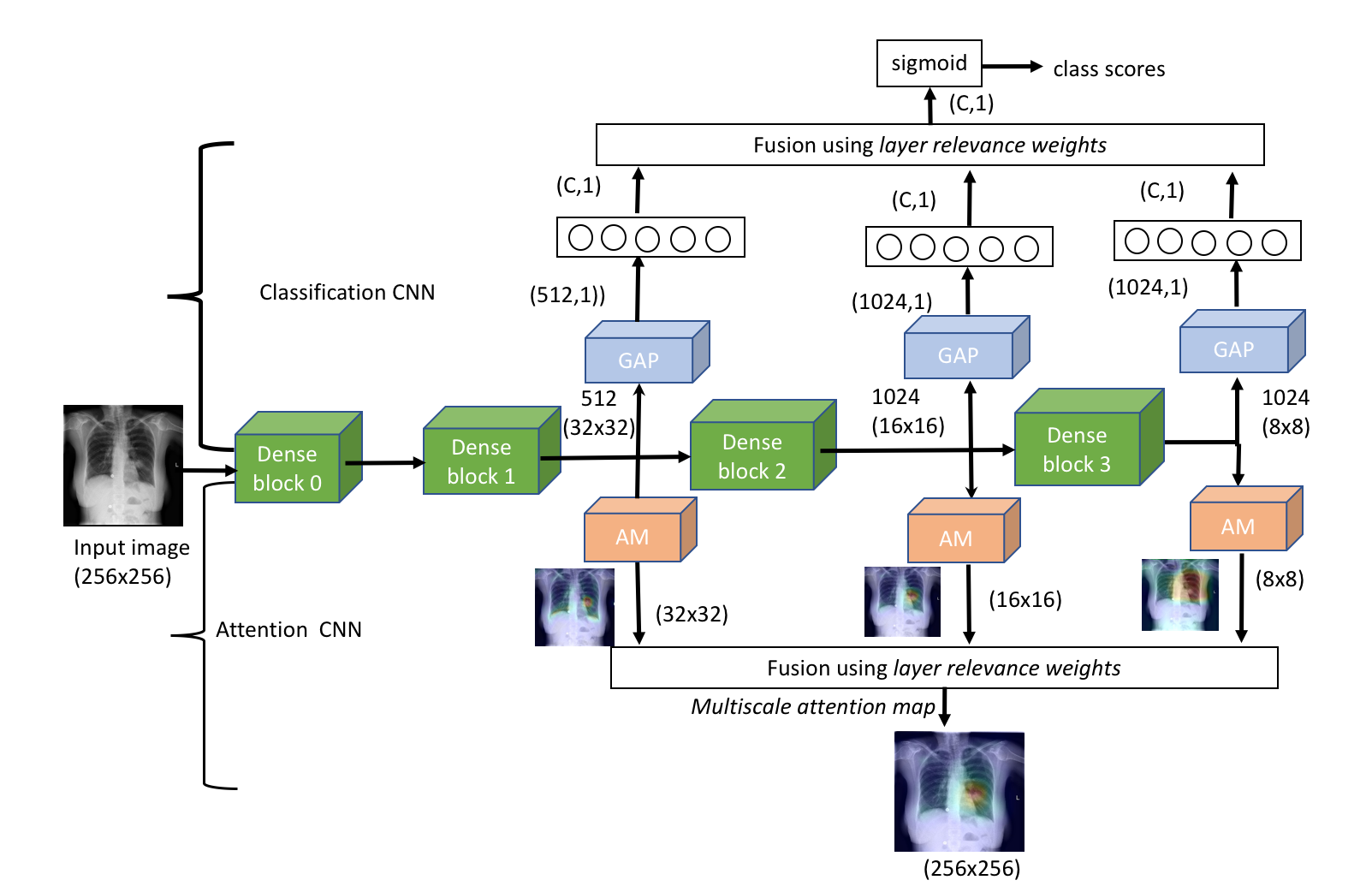} 
		\caption{Proposed chest X-ray  pathology localization  framework: the classification CNN (C-CNN)   combines the  intermediate feature maps using learned \emph{layer relevance weights}. The attention CNN (A-CNN) computes the \emph{multiscale attention map}   using  the feature maps and  the learned \emph{ layer relevance weights}.} 	\label{fig:block_diagram}
	\end{figure}

	Our proposed method learns pathology localization from image-level supervision where training images are known to contain the instance of pathology class but their locations in the image are unknown. As shown in Figure \ref{fig:block_diagram}, the designed network uses the base network of DenseNet blocks.  The network switches between 2 modes during training and test phases; 1) Classification CNN (C-CNN) and 2) Attention CNN (A-CNN), respectively. We first train the C-CNN by enforcing both shallower and deeper convolutional layers to contribute to the overall classification of pathologies. We introduce a class specific \emph{layer relevance weights} to combines the feature maps from these layers and the classification is performed by only the \emph{convex combination} of the responses from the feature maps. In prediction phase, A-CNN combines the convolutional feature maps from individual layers using the learned \emph{layer relevance weights} to obtain the \emph{multiscale attention map}. The proposed \emph{multiscale attention map} is robust against pathology size as it encapsulates the feature maps from both coarse and fine layers.

	\subsection{Classification-CNN}
	We present a general framework for multiscale feature learning for localization. The network architecture we chose is 121 layers Densenet \cite{Huang2017}. It consists of four dense blocks where  each block consists  of several convolution layers. Each layer in the \emph{dense block} is connected to all the preceding layers by iterative concatenation of previous \emph{feature maps}. This allows all layers to access \emph{feature maps} from their preceding layers which encourages heavy feature reuse. The feature maps at end of each block are down-sampled and passed to the next block and the global average pooling response of the feature maps from the final block  are connected to a densely connected network to obtain the classification scores. 
	
	The issue with this base architecture is that its ability to localize the small pathologies is compromised due to successive down-sampling of feature maps. We modify this architecture to leverage the intermediate feature maps. In order to do so, we plug in the \emph{ global average pooling} (GAP) operator at the end of each dense block as shown in Fig \ref{fig:block_diagram}. The pooling operation provides structural regularization to the network \cite{Zhou2016}, hence it facilitates learning of meaningful feature maps.  
	
	Let $F_b$ be the feature maps from block $b$. The dimension of   $F_b$  is dependent on the number of convolution layers on the block. As shown in Figure \ref{fig:block_diagram}, $F_1 \in \mathbb{R}^{512\times32\times32}$, $F_2 \in \mathbb{R}^{1024\times16\times16}$ and $F_3 \in \mathbb{R}^{1024\times8\times8}$. We apply  \emph{global average pooling}  operation to $F_1$, $F_2$ and $F_3$  to obtain the pooled feature maps of dimension $\mathbb{R}^{512\times1\times1}$, $\mathbb{R}^{1024\times1\times1}$ and $\mathbb{R}^{1024\times1\times1}$. These pooled feature maps are flattened and passed through separate fully connected layer to obtain the block-specific $C$ dimensional \emph{logits} vectors i.e.,  $ L_b =[l_b^1,\cdots,l_b^C], b=1,\cdots,B$. The \emph{logit} response from all the layers have same dimension (equal to the number of category for classification) and now can be combined using class specific convex combination to obtain the probability score for the class $p_c$.
	\begin{equation}\label{eq:logit_fusion}
	p_c=\sigma (\sum_{b=1}^B w_b^c \times l_b^c)
	\end{equation}
where $\sigma$ is \emph{sigmoid} function; $w_b^c$ is the \emph{layer relevance  weight} assigned to the $b^{th}$ block to predict the $c^{th}$ class  and follows the  \emph{convex} weight constraint  as described below.

\subsection{Class aware training of convolutional features} 
\label{sec:training}
We are given the training data set $\left\{ I_{n}, \mathbf{y}_{n}\right\} _{n=1}^{N}$ where $I_{n}$ is the input image and $\mathbf{y}_{n} $ is a label vector. For brevity, we drop the subscript $n$. The label vector is given by $ \mathbf{y}=[y_1,\cdots,y_C] ;y_{i}\in\left\{ 0,1\right\} $ indicates the presence of the $c^{th}$ pathology class in the image and $C$ is a number of pathology classes. Let $W$ be the weights of the C-CNN including the \emph{layer relevance weights} $w_b^c$.  We initialize the layer relevance weights  $w_b^c$  to  $1/B$ and the remaining network weights with Xavier initialization. We then optimize the network weights $W$  by minimizing the class-balanced cross entropy loss with a \emph{convex} weight constraint: 
	
	\begin{equation*}
	\begin{aligned}
	& \underset{W}{\text{minimize}}
	& & L(W)=\sum_{c=1}^C (-\beta_c\sum_{y_c=1} \text{log }p_c- \sum_{y_c=0} (1-\beta_c) \text{log } (1-p_c)) \\
	& \text{subject to}
	& & w_b^c \ge 0 \text{ and } \sum_{b=1}^B w_b^c = 1 \; c = 1, \ldots, C.
	\end{aligned}
	\end{equation*}
where $\beta_c$ is a balancing factor which denotes the percentage of '0' samples in the ground truth i.e, $\beta_c= \frac{|y_c=0|}{|y_c|}$. The balancing factor is used to mitigate the effect of large number of '0' samples. The \emph{convex}  weights constraint enables the  probabilistic combination  of \emph{logits} from each block as shown by Equation \ref{eq:logit_fusion}. As a result, the learned weights encodes the relevance of each block in classifying a given pathology.  The proposed network is trained using mini-batch gradient descent and the Adam optimizer with momentum and a batch size of 32. The learning rate is set to $10^{-3}$ which is decreased by a factor of 0.1 whenever the \emph{validation loss} reaches plateau.

\subsection{Pathology Localization by Attention CNN (A-CNN)}
\label{sec:localization}

C-CNN presented above enables individual blocks in the network to learn  relevant feature maps   with respect to each other. A-CNN uses the weights learned from C-CNN to compute the \emph{multiscale attention map} of pathologies (Figure \ref{fig:block_diagram}).   First, the attention map for each block are computed using the CAM technique. The \emph{multiscale attention map} is then obtained using convex combination of attention map at each block. The CAM at each block can be computed using the weighted average of the feature maps of the block using the learned fully connected weights.  Let $V_j^c(b)$ denote  the sampled FC weights  which connects $j^{th}$ feature map from $b^{th}$ block to the $c^{th}$ class.  The  attention map of the $c^{th}$ class at the $b^{th}$ dense-block can be computed as: 
	\begin{equation}\label{eq:cam}
	S_b^c=\sum_{j=1}^{N_f(b)} V_j^c(b) F_j^b
	\end{equation}
where $N_f(b)$ is the number of feature maps at the $b$-th block.  The  \emph{multiscale attention map} for each class can now be obtained as a convex combination of intermediate attention maps:
	
	\begin{equation}\label{eq:cam2}
	S_c=\sum_{i=1}^B w_b^c R(S_b^c), c = {1, \ldots, C}
	\end{equation}
where $R()$ is a function that takes an intermediate attention map $S_b^c$ and resizes it to the same spatial resolution as the input image and $w_b^c$ is the layer relevance weights of A-CNN.  The resulting attention map $S_c$ encapsulates the feature maps  from all the blocks through class specific  probabilistic  combination of attention maps from individual block using the weights learned during training phase.

\section{Experiments}
	
We use the ChestX-ray14 dataset \cite{Wang2017}, which is the largest collection of public chest X-ray dataset by far. It  consists of 112,120  frontal-view  chest  X-ray  images  of 30,805 unique patients. Each image is labeled with one or  more  types  of  $14 $ common  thorax  diseases. Also, for a subset of $983$ images,  bounding box annotations of $8$ pathologies are provided for the evaluation of weakly supervised localization methods. 

We randomly split the dataset into  $70\% $ for training,  $10\%$ for validation  and $20\%$ for test using patient id  to ensure there is no patient overlap. We also make sure that the images with bounding box annotations falls only in the test set. The images are downscaled to the size of $256 \times 256$ before feeding to the network. The classification network is then trained using the method described in Section \ref{sec:training}. During test phase, we compute the multiscale attention map of each pathology using the method described in Section \ref{sec:localization}. The attention map of the pathology gives approximate spatial location of the pathology in the input image. The attention map is converted to the bounding box by simple thresholding of the attention map and enclosing the resulting masks with the rectangles. We then evaluate the performance of the  predicted bounding boxes against the ground truth bounding boxes. 

We compare our method with the baseline Resnet-CAM(RN-CAM)  \cite{Wang2017} and Densenet-CAM (DN-CAM).  Both networks use only the feature maps from the deepest convolution layers to localize the pathologies. We use intersection over union (IOU)  ratio between the predicted and ground truth bounding boxes as the detection criteria. We consider positive detection when IOU is greater than a  given threshold value. IOU is commonly used measure in evaluation of object detection \cite{Girshick2015,Wang2017}. We evaluate our localization method (A-CNN) for two different thresholds of 0.3 and 0.5.

\begin{table}[tbh]
	\caption{Localization accuracy and average false positive (AFP)  of our A-CNN compared to the state of the art weakly supervised localization methods RN-CAM \cite{Wang2017} and DN-CAM.}
	\label{tab:res}
	\centering{}%
	\begin{tabular}{|c|c|c|c|c|c|c|}
		\hline 
		& \multicolumn{6}{c|}{Localization Accuracy / (AFP)}\tabularnewline
		\hline 
		& \multicolumn{3}{c|}{T(IOU)$>$0.3} & \multicolumn{3}{c|}{T(IOU) $>$0.5}\tabularnewline
		\hline 
     	Pathology & RN-CAM  & DN-CAM & A-CNN & RN-CAM  & DN-CAM & A-CNN \tabularnewline
		\hline 
		Atelectasis & $0.24$ (1.0) & $0.17$ (0.9) & $\mathbf{0.30}$ (0.8) & $0.05$ (1.0) & $0.01$ (1.1) & \textbf{$\mathbf{0.1}$ }(0.9) \tabularnewline
		\hline 
		Cardiomegaly & $0.45$ (0.7) & $\mathbf{0.86}$ (0.2) & $0.84$ (0.2) & $0.17$ (0.8) & $\mathbf{0.51}$ (0.4) & \textbf{$0.47$ }(0.5) \tabularnewline
		\hline 
		Effusion & $\mathbf{0.3}$ (0.9) & $0.14$ (0.9) & $0.25$ (0.8) & $\mathbf{0.11}$ (0.9) & $0.01$ (0.9) & \textbf{$0.03$ }(0.9) \tabularnewline 
		\hline 
		Infiltration & $0.27$ (0.7) & $0.21$ (0.6) & $\mathbf{0.32}$ (0.7) & $0.06$ (0.7) & $0.10$ (0.7) & \textbf{$\mathbf{0.12}$ }(0.8) \tabularnewline
		\hline 
		Mass & $0.15$ (0.7) & $0.17$ (0.6) & $\mathbf{0.48}$ (0.5) & $0.01$ (0.7) & $0.05$ (0.8)  & \textbf{$\mathbf{0.2}$ }(0.8) \tabularnewline
		\hline 
		Nodule & $0.03$ (0.6) & $0.0$3 (0.6) & $\mathbf{0.27}$ (0.7) & $0.01$ (0.6) & $0.01$ (0.8) & \textbf{$\mathbf{0.11}$ }(0.7) \tabularnewline
		\hline 
		Pneumonia & $0.16$ (1.1) & $0.20$ (0.9) & $\mathbf{0.4}$ (0.8) & $0.03$ (1.1) & $0.09$ (0.9) & \textbf{$\mathbf{0.10}$ }(0.8) \tabularnewline
		\hline 
		Pneumothorax & $0.13$ (0.5)  & $0.11$ (0.6) & $\mathbf{0.31}$ (0.8) & $0.03$ (0.5)  & $0.04$ (0.6) & \textbf{$\mathbf{0.08}$}(0.8 \tabularnewline
		\hline 
	\end{tabular}
\end{table}

\begin{figure}[t]
		\centering \includegraphics[scale=0.22]{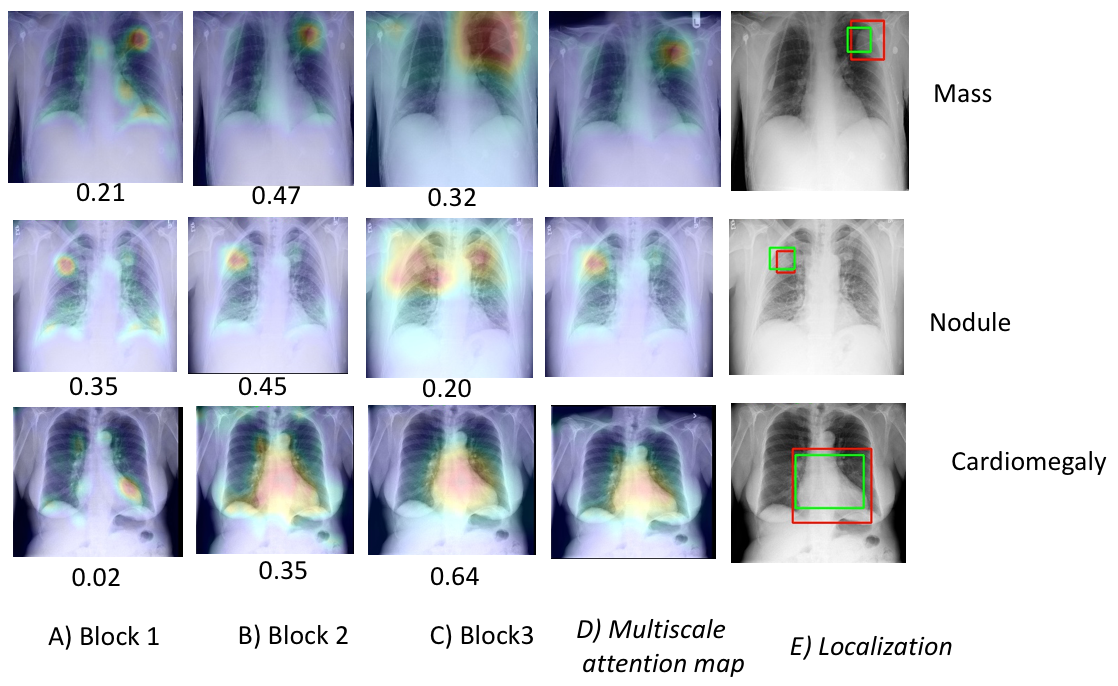} 
		\caption{Localization examples of  few pathologies  using the proposed \emph{multiscale attention map}.  The layer relevance weights are shown below the attention maps.}	\label{fig:result_pathologies}
	\end{figure}

	Table \ref{tab:res} compares the pathology detection accuracy and average false positive (AFP) of our proposed method compared with RN-CAM  \cite{Wang2017} and with the  DN-CAM  both without using intermediate feature maps i.e., they use feature maps from only the deepest layer for localization.  The proposed method outperforms Resnet for all pathologies except for comparable performance for effusion. Our method also gives improved localization accuracy compared to Densnet for every pathology except cardiomegaly where the performance is comparable. Particularly, the proposed method gives notably improved accuracy for most challenging cases of small pathologies such as \emph{mass}  $0.48$ and \emph{nodule} $0.27$ in comparison to both RN-CAM and DN-CAM. Figure \ref{fig:result_pathologies} shows the examples of localization and \emph{mutiscale attention map} produced by our proposed method  along with the attention map produced at individual dense block of the network and corresponding \emph{layer relevance weight}. The proposed \emph{multiscale attention map} is obtained by class specific probabilistic combination of  feature maps from both coarser and deeper blocks, therefore can capture small and large pathologies using a single network.  It can be observed that the \emph{layer relevance weights} put emphasis on shallower blocks for smaller pathologies and larger weights on deeper blocks  for larger pathology classes.  
		
	
	\section{Conclusion} 
	In this paper, we propose a novel weakly supervised method based on class aware multiscale convolutional feature learning to localize chest pathologies in X-ray images. The classification CNN learns to classify pathology responses from the intermediate feature maps along with the class specific layer relevance weights for coarser and deeper layers. In the test phase, the learned layer relevance weights are used to perform the probabilistic combination of the intermediate feature maps from the CNN to obtain the multiscale attention map for pathology localization. Experimental results demonstrate that the proposed weakly supervised method significantly improves the localization performance of small sized pathologies, such as \emph{nodule} and \emph{mass} which are particularly challenging to locate in X-ray scans, while giving comparable performance for bigger pathologies such as effusion and cardiomegaly. The proposed method has a very practical use in multitude of real-world problems where the availability of quality annotation for localization in very scarce and pathologies of different sizes could be present in the same image. These conditions frequently occur in medical data such as X-ray images. Automated systems, powered by the proposed method, have a great potential to enhance the effectiveness of a computer aided diagnosis system by increasing the rate of incidental findings in routine check-ups.

	\bibliographystyle{splncs03}
	\bibliography{refs1}

\begin{thebibliography}{10}
\providecommand{\url}[1]{\texttt{#1}}
\providecommand{\urlprefix}{URL }

\bibitem{Uri2011}
Avni, U., Greenspan, H., Goldberger, J.: X-ray categorization and spatial
  localization of chest pathologies. In: Fichtinger, G., Martel, A., Peters, T.
  (eds.) Proc. MICCAI. pp. 199--206 (2011)

\bibitem{Jia2017}
Ding, J., Li, A., Hu, Z., Wang, L.: Accurate pulmonary nodule detection in
  computed tomography images using deep convolutional neural networks. In:
  Descoteaux, M., Maier-Hein, L., Franz, A., Jannin, P., Collins, D.L.,
  Duchesne, S. (eds.) Proc. MICCAI. pp. 559--567 (2017)

\bibitem{Girshick2015}
Girshick, R.: Fast {R-CNN}. In: IEEE ICCV. pp. 1440--1448 (Dec 2015)

\bibitem{Huang2017}
Huang, G., Liu, Z., v.~d. Maaten, L., Weinberger, K.Q.: Densely connected
  convolutional networks. In: IEEE Conference on CVPR. pp. 2261--2269 (July
  2017)

\bibitem{Sangheum2016}
Hwang, S., Kim, H.E.: Self-transfer learning for weakly supervised lesion
  localization. In: Ourselin, S., Joskowicz, L., Sabuncu, M.R., Unal, G.,
  Wells, W. (eds.) Proc. MICCAI. pp. 239--246 (2016)

\bibitem{Liao2013}
Liao, S., Gao, Y., Lian, J., Shen, D.: Sparse patch-based label propagation for
  accurate prostate localization in {CT} images. IEEE Transactions on Medical
  Imaging  32(2),  419--434 (2013)

\bibitem{Oquab2015}
Oquab, M., Bottou, L., Laptev, I., Sivic, J.: Is object localization for free?
  - weakly-supervised learning with convolutional neural networks. In: EEE
  Conference on CVPR. pp. 685--694 (2015)

\bibitem{Pesce2017}
Pesce, E., Ypsilantis, P., Withey, S., Bakewell, R., Goh, V., Montana, G.:
  Learning to detect chest radiographs containing lung nodules using visual
  attention networks. CoRR  abs/1712.00996 (2017)

\bibitem{Olaf2015}
Ronneberger, O., Fischer, P., Brox, T.: U-net: Convolutional networks for
  biomedical image segmentation. In: Navab, N., Hornegger, J., Wells, W.M.,
  Frangi, A.F. (eds.) Proc. MICCAI. pp. 234--241 (2015)

\bibitem{Sedai2017}
Sedai, S., Tennakoon, R., Roy, P., Cao, K., Garnavi, R.: Multi-stage
  segmentation of the fovea in retinal fundus images using fully convolutional
  neural networks. In: ISBI. pp. 1083--1086 (April 2017)

\bibitem{seg2017}
Sedai, S., Mahapatra, D., Hewavitharanage, S., Maetschke, S., Garnavi, R.:
  Semi-supervised segmentation of optic cup in retinal fundus images using
  variational autoencoder. In: MICCAI. pp. 75--82 (2017)

\bibitem{Wang2017}
Wang, X., Peng, Y., Lu, L., Lu, Z., Bagheri, M., Summers, R.M.: Chestx-ray8:
  Hospital-scale chest x-ray database and benchmarks on weakly-supervised
  classification and localization of common thorax diseases. CoRR
  abs/1705.02315 (2017)

\bibitem{XU2014591}
Xu, Y., Zhu, J.Y., Chang, E.I.C., Lai, M., Tu, Z.: Weakly supervised
  histopathology cancer image segmentation and classification. Medical Image
  Analysis  18(3),  591 -- 604 (2014)

\bibitem{Zhou2016}
Zhou, B., Khosla, A., Lapedriza, A., Oliva, A., Torralba, A.: Learning deep
  features for discriminative localization. In: IEEE Conference on CVPR. pp.
  2921--2929 (June 2016)

\end{thebibliography}

\end{document}